\documentclass[11pt,a4paper]{article}

\usepackage{times,latexsym}
\usepackage{url}
\usepackage[T1]{fontenc}

\usepackage[acceptedWithA]{tacl2021}

\usepackage[utf8]{inputenc}

\usepackage{subcaption}
\usepackage{microtype}
\usepackage{booktabs} 
\usepackage{multirow}
\usepackage{hyperref}
\usepackage{xcolor}
\usepackage{url}
\usepackage{caption}
\usepackage{tabularx}
\usepackage{multicol}
\usepackage{floatrow}
\usepackage{microtype}
\usepackage{amsbsy}
\usepackage{amsmath}
\usepackage{xspace}
\usepackage{inconsolata}
\usepackage{amssymb}
\usepackage{adjustbox}
\usepackage{textcomp}
\usepackage{color, colortbl}
\usepackage{arydshln}
\usepackage{pifont}
\usepackage{arydshln}
\usepackage{euflag}

\newcommand{\dataset}{{\textsc{Multi$^3$WOZ}}\xspace}

\newcommand{\eng}{{\textsc{eng}}\xspace}
\newcommand{\ara}{{\textsc{ara}}\xspace}
\newcommand{\fra}{{\textsc{fra}}\xspace}
\newcommand{\tur}{{\textsc{tur}}\xspace}

\newcommand{\rparagraph}[1]{\vspace{1mm}\noindent\textbf{#1.}}
\newcommand{\rrparagraph}[1]{\vspace{0.5mm}\noindent\textit{#1:}}

\definecolor{Gray}{gray}{0.92}
\definecolor{racing-green}{rgb}{0.0, 0.8, 0.6}
\definecolor{awesome-red}{rgb}{1.0, 0.13, 0.32}
\newcolumntype{Y}{>{\centering\arraybackslash}X}

\newcommand{\tod}{{\textsc{ToD}}\xspace}
\newcommand{\bluecheck}{{\color{racing-green}{\pmb{\checkmark}}}}
\newcommand{\xmark}{\color{awesome-red}\ding{55}}%

\usepackage{todonotes}
\makeatletter
\newcommand*\iftodonotes{\if@todonotes@disabled\expandafter\@secondoftwo\else\expandafter\@firstoftwo\fi}
\makeatother

\usepackage[inline]{enumitem}
\usepackage{bbm}

\definecolor{wozblue}{HTML}{BED3F3}
\definecolor{wozgreen}{HTML}{BEDADC}
\definecolor{wozverybleu}{HTML}{1273de}
\definecolor{wozveryred}{HTML}{b80000}
\definecolor{wozverypurple}{HTML}{5300eb}

\newif\iftaclinstructions
\taclinstructionsfalse %
\iftaclinstructions
\renewcommand{\confidential}{}
\renewcommand{\anonsubtext}{(No author info supplied here, for consistency with
TACL-submission anonymization requirements)}
\newcommand{\instr}
\fi

\author{
  Songbo Hu$^{1}$\thanks{~~Equal contribution.}~~~ 
  Han Zhou$^{1}$\footnotemark[1]~~~
  Mete Hergul$^{1}$\\
  \textbf{Milan Gritta$^{2}$~~~
  Guchun Zhang$^{2}$~~~
  Ignacio Iacobacci$^{2}$}\\
  \textbf{Ivan Vuli\'{c}$^{1}$\thanks{~~Equal senior contribution.}~~~
  Anna Korhonen$^{1}$\footnotemark[2]}
  \\
  $^{1}$Language Technology Lab, University of Cambridge, UK
  \\
  $^{2}$Huawei Noah’s Ark Lab, London, UK
  \\
  $^{1}$\texttt{\{sh2091,hz416,mh2071,iv250,alk23\}@cam.ac.uk} \\
  $^{2}$\texttt{\{milan.gritta,guchun.zhang,ignacio.iacobacci\}@huawei.com}
}

\date{}

\title{\dataset: A Multilingual, Multi-Domain, Multi-Parallel Dataset for Training and Evaluating Culturally Adapted Task-Oriented Dialog Systems}

\begin{document}

\maketitle
\begin{abstract}
Creating high-quality annotated data for task-oriented dialog (\tod) is known to be notoriously difficult, and the challenges are amplified when the goal is to create equitable, culturally adapted, and large-scale \tod datasets for multiple languages. Therefore, the current datasets are still very scarce and suffer from limitations such as translation-based non-native dialogs with translation artefacts, small scale, or lack of cultural adaptation, among others. In this work, we first take stock of the current landscape of multilingual \tod datasets, offering a systematic overview of their properties and limitations. Aiming to reduce all the detected limitations, we then introduce \textbf{\dataset}, a novel multilingual, multi-domain, multi-parallel \tod dataset. It is large-scale and offers culturally adapted dialogs in 4 languages to enable training and evaluation of multilingual and cross-lingual \tod systems. We describe a complex bottom-up data collection process that yielded the final dataset, and offer the first sets of baseline scores across different \tod-related tasks for future reference, also highlighting its challenging nature.
\end{abstract}

\section{Introduction and Motivation}
\label{s:introduction}

Task-oriented dialog (\tod), where a human user engages in a conversation with a system agent
with the aim of completing a concrete task, is one of the central objectives, hallmarks, and applications of machine intelligence \cite[\textit{inter alia}]{Gupta:2006,Tur:2010,Young:2010}. \tod technology has been proven useful across  a wide spectrum of application sectors such as hospitality industry \cite{Henderson:2014dstc3,Henderson:2019poly}, healthcare \cite{Laranjo:2018healthcare}, online shopping \cite{Yan:2017aaai}, banking \cite{Altinok:2018arxiv}, and travel \cite{Raux:2005letsgo,el-asri-etal-2017-frames}, among others.

Wider developments in \tod have been hampered by the two conflicting requirements: \textbf{1)} large-scale in-domain datasets are crucially required in order to unlock the potential of deep learning-based \tod components and systems to handle complex dialog patterns \cite{Budzianowski:2018multiwoz,Lin:2021bitod}; at the same time \textbf{2)} data collection for \tod is known to be notoriously difficult as it is extremely time-consuming, expensive, and requires expert and domain knowledge \cite{Shah:2018naacl,Larson:2022arxiv}. Put simply, the creation of \tod datasets for new domains and languages incurs significantly higher time and budget costs than for most other NLP tasks \cite{Casanueva:2022nlupp}. Consequently, the progress in \tod until recently has been limited only to a small number of high-resource languages such as English and Chinese \cite{Razumovskaia:2022survey}.

Recent work has recognized the need to expand the reach of multilingual \tod technology to more languages via collecting multilingual \tod data \cite{Razumovskaia:2022survey}. Yet, as discussed in more detail later in \S\ref{s:rw}, all the currently available multilingual \tod datasets suffer from one or several serious limitations: (i) the predominant reliance on translation-based data creation that introduces issues with `translationese' and artificial performance inflation \cite{xu-etal-2020-end,Zuo:2021allwoz}; (ii) lack of \textit{cultural adaptation} also results in artificial dialogs that are not localized nor adapted to real-world data and to cultural specificities of each target language and culture; (iii) small scale and lack of sufficient training data prevents truly equitable multilingual development and in-depth comparative cross-language analyses \cite{ding-etal-2022-globalwoz,hung-etal-2022-multi2woz}; (iv) lack of coherent and multi-parallel dialogs in all the represented languages, which are typically not created and corrected by native speakers, hinders meaningful cross-language comparisons and analyses \cite{ding-etal-2022-globalwoz}; (v) some datasets focus on a single component of a full \tod system, typically Natural Language Understanding (NLU), which prevents training and evaluation of other crucial tasks such as Dialog State Tracking (DST), or Natural Language Generation (NLG) in multilingual and transfer setups.

In this work, we address all the aforementioned limitations of current multilingual \tod datasets and present a large-scale data collection process that resulted in a novel large-scale multilingual dataset for \tod: \textbf{\dataset}. The departure point of our data collection is the established \textit{multi-domain} English MultiWOZ dataset \cite{Budzianowski:2018multiwoz}, that is, its cleaned version 2.3 in particular \cite{Han:2021multiwoz23}. \dataset is then created via adapting a recent \textit{bottom-up outline-based} approach of \newcite{Majewska:2023cod} which bypasses (the issues of) the translation-based design and discerns between language-agnostic \textit{abstract dialog schemata} (i.e., \textit{outlines}) and adapted, language-specific \textit{surface realizations} of the underlying schemata (i.e, the actual user and system \textit{utterances}). We validate the usefulness and feasibility of the outline-based approach to multilingual \tod data creation for the first time on a large scale, and prove its feasibility for such large-scale endeavors: the dataset contains a total of 494,116 dialog turns created manually by human subjects.

Guided by the need to tackle the present limitations, \dataset is the first multilingual \tod dataset with the following crucial properties; see also Table~\ref{tab:summary_dataset} for an overview.
First, \dataset is \textit{large-scale} with the equal number of training (7,440 dialogs per language), development (860), and test dialogs (860) offered in 4 different languages: English, Arabic, French, and Turkish. It is more versatile than all prior multilingual \tod datasets as it allows for training and evaluation in monolingual, multilingual, and cross-lingual setups, and in zero-shot, few-shot, and `many'-shot \textit{cross-lingual} and \textit{cross-domain} transfer scenarios. Second, \dataset offers \textit{multi-parallel} dialogs, conveying comparable information over exactly the same conversational flows across all four languages. This property allows for cross-language studies and comparative analyses. Third, \dataset enables \textit{both} (monolingual and multilingual) training and evaluation over different constituent \tod tasks such as NLU (intent detection and slot filling), DST, NLG, as well as full-fledged end-to-end (E2E) learning. Fourth, \dataset is \textit{localized} and \textit{culturally adapted} to the actual existing entities from the cultures in which the target languages are spoken. Finally, created in a bottom-up fashion by native speakers of the target languages, hence \textit{linguistically adapted} to the target language, it offers natural and native dialogs in all target languages, avoiding `translationese' and preventing over-inflation of transfer performance~\cite{Majewska:2023cod}.

Furthermore, to guide future research, we set reference scores across different \tod tasks in all the languages of \dataset, running a representative set of standard baselines in each relevant \tod task. The results clearly indicate the challenging nature of the dataset; we also outline the differences in performance across different languages.

\begin{table*}[t!]
\centering
\def\arraystretch{0.75}
{\scriptsize 
\resizebox{\textwidth}{!}{
\begin{tabular}{l cccc cccc}
\toprule
\rowcolor{Gray}
\textbf{Dataset (Reference)} & \# Langs    & \# Domains     & \# Train & \# Test & No Translation? & Culturally Adapted? & Coherent? & Multi-P? \\
\cmidrule(lr){2-5} \cmidrule(lr){6-9}
WOZ 2.0~\cite{mrksic-etal-2017-semantic} & {3} & {1} & {600} & {400} & {\xmark} & {\xmark} & {\bluecheck} & {\bluecheck} \\
BiToD~\cite{Lin:2021bitod} & {2} & {5} & {2,894} & {451} & {\bluecheck} & {\bluecheck} & {\bluecheck} & {\xmark} \\
AllWOZ~\cite{Zuo:2021allwoz} & {8} & {5} & {40} & {50} & {\xmark} & {\xmark} & {\bluecheck} & {\bluecheck} \\
GlobalWOZ~\cite{ding-etal-2022-globalwoz} & {21} & {7} & {0 (8,437)}$^{*}$ & {500 (1,000)}$^{*}$ & {\xmark} & {\bluecheck} & {\xmark} & {\xmark} \\
Multi$^2$WOZ~\cite{hung-etal-2022-multi2woz} & {5} & {7} & {0} & {1,000} & {\xmark} & {\xmark} & {\bluecheck} & {\bluecheck} \\
\cmidrule(lr){2-5} \cmidrule(lr){6-9}
\textbf{\dataset} (this work) & {4} & {7} & {7,440} & {860} & {\bluecheck} & {\bluecheck} & {\bluecheck} & {\bluecheck} \\
\bottomrule
\end{tabular}
}
}%
\caption{Summary of multilingual \tod datasets that support multiple languages and \tod tasks (including E2E learning), with more details concerning each dimension of comparison available in \S\ref{s:rw}. For clarity, we do not show (i) monolingual \tod datasets constructed for languages other than English, we refer the reader to the survey of \newcite{Razumovskaia:2022survey} for a comprehensive overview; as well as (ii) the body of multilingual \tod datasets that focus solely on NLU for \tod (see \S\ref{s:rw}). \textbf{\# Langs} refers to the total number of languages in each dataset, including English. \textbf{\# Train} and \textbf{\# Test} refer to the average number of human-created or human-curated dialogs \textit{per each language} in the respective portions of each dataset. \textbf{Multi-P} refers to multi-parallelism of dialogs in the dataset. $(*)$ GlobalWOZ releases training data created automatically by an English-target NMT system, without any human curation nor post-processing, and manually curates only a portion of 500 dialogs from the target language test sets (see \S\ref{s:rw} for more details).}
\label{tab:summary_dataset}
\end{table*}

\section{\dataset versus Limitations of Current Multilingual \tod Datasets}
\label{s:rw}
We now delve deeper into the main benefits of \dataset, characterizing how its key properties make it a unique \tod resource. The summary and statistics of the most relevant prior work are provided in Table~\ref{tab:summary_dataset}. Building upon this table, we discuss those datasets along with other related work in what follows, focusing on the five desirable properties of \dataset and how these counteract the detected main limitations of other datasets.

\vspace{0.5mm}
\noindent \textbf{P1. Supporting Multiple Languages and \tod Tasks.} 
There has been a surge of interest in the creation of multilingual \tod datasets, aiming to mitigate the language resource gap in multilingual NLP \cite{ponti-etal-2019-modeling,joshi-etal-2020-state}. Despite the effort, the gap is still much more pronounced for dialog tasks and data than for some other NLP tasks such as NLI \cite{conneau-etal-2018-xnli,Ebrahimi:2022americasnli} or NER~\cite{adelani-etal-2021-masakhaner}, also due to its increased time demands and cost of annotation.\footnote{For instance, the creation of the validation and test sets of the XCOPA dataset requires a total time ranging from 12 to 20 hours per language \cite{ponti-etal-2020-xcopa}. In contrast, the creation of the validation and testing sets for each individual language in \dataset requires over 300 hours of effort. Even when considering the annotation cost per sentence (utterance), which amounts to approximately \$0.17 per utterance, the cost is notably higher than the per sentence annotation cost for NER (\$0.06 as reported by~\citet{Bontcheva2017}) and NLI (\$0.01015 per instance as reported by~\citet{marelli-etal-2014-sick}).}  Further, the majority of multilingual \tod datasets focused only on two standard NLU tasks (i.e., intent detection and slot labeling), again due to the high cost and specific challenges posed by collecting full dialog data \cite{Budzianowski:2018multiwoz}. The first wave of such NLU datasets were built upon the single-domain English ATIS dataset \cite{hemphill-etal-1990-atis}, extending it to 10 languages via human translation \cite{Upadhyay:2018icassp,xu-etal-2020-end,Dao:2021interspeech}. More recent NLU datasets cover multiple domains and wider linguistic typology and geography \cite{schuster-etal-2019-cross-lingual,FitzGerald:2022arxiv,Moghe:2022multi3nlu,Majewska:2023cod}. However, current NLU datasets (i) still support only the two NLU tasks, and (ii) provide utterances `in isolation' (i.e., out of the context of the full dialog which facilitates their multilingual construction). Further, (iii) some datasets do not provide any training data and are useful only for evaluation of (zero-shot) cross-lingual transfer; (iv) all the datasets except that of \citet{Majewska:2023cod} and the concurrent work of \newcite{presto} were constructed via translation from the source English datasets.

Monolingual `end-to-end' \tod datasets, which support NLU as well as other \tod tasks (i.e., modeling and evaluation of the full \tod pipeline), have been created only for particular high-resource languages. MultiWOZ \cite{Budzianowski:2018multiwoz} and Taskmaster \cite{byrne-etal-2019-taskmaster} are two large-scale multi-domain English datasets spanning 7 and 6 domains, respectively, containing both single-domain and multi-domain dialogs. Inspired by MultiWOZ, monolingual RisaWOZ \cite{quan-etal-2020-risawoz} and CrossWOZ \cite{zhu-etal-2020-crosswoz} datasets have been created for Chinese.
 Crucially, multilingual multi-domain \tod datasets that support full \tod modeling are still scarce, see Table~\ref{tab:summary_dataset}, and they all come with some core limitations, as discussed next.

\vspace{0.5mm}
\noindent \textbf{P2. Avoiding Translation-Based Design.} The majority of datasets have been obtained via \textit{manual or semi-automatic translation} (e.g., via post-editing MT output - PEMT) of an English source dataset \cite{Zuo:2021allwoz,ding-etal-2022-globalwoz,hung-etal-2022-multi2woz}. The translation-based approach is cost-efficient and can natively yield data which is comparable across languages, but results in (i) undesired `translationese' effects \cite{Artetxe:2020emnlp}, (ii) lacks dialog naturalness \cite{ding-etal-2022-globalwoz}, and (iii) typically leads to overinflated and thus misleading performance of \tod systems. For instance, \citet{Majewska:2023cod} empirically validate that cross-lingual transfer performance substantially increases when exactly the same dialogs are obtained via automatic or manual translation rather than via a bottom-up approach relying on native speakers of the target languages. 

Unlike prior work (i.e., all datasets from Table~\ref{tab:summary_dataset} except BiToD), the honed outline-based construction of \dataset (see \S\ref{s:dataset} later) avoids all the negative implications of translation, while maintaining cost efficiency (and thus enabling its large scale), supporting cultural adaptation, and enabling coherence and multi-parallelism.

\vspace{0.5mm}
\noindent \textbf{P3. Dataset Scale and Large-Scale Training.}
\dataset offers a substantially larger number of dialogs for training than any previous multilingual `full \tod' dataset, and it treats the four supported languages in an equitable way: i.e., it provides the same set of manually (bottom-up) constructed dialogs for training, development, and testing in each language. Previous work (Multi$^2$WOZ, AllWOZ, GlobalWOZ) targeted the creation of test data only, for evaluating cross-lingual transfer scenarios. These datasets come (i) without providing any training data at all (Multi$^2$WOZ), or (ii) with a very small set of post edited MT-obtained dialogs (AllWOZ),\footnote{The tiny size of AllWOZ is even more problematic at the level of single domains, e.g., it contains only 13 dialogs for the \textit{Taxi} domain, hindering any generalisable evaluations.} or (iii) with automatically created MT-based training data only (GlobalWOZ). The only exception is BiToD \cite{Lin:2021bitod}, but it spans only two, highest-resourced languages, a smaller number of domains, and has approximately three times fewer training data than \dataset. For instance, \dataset contains almost 124,000 turns \textit{per each represented language} ($\sim$98,000/12,500/12,500), with a total of 494,116 turns; for comparison, the \textit{total} number of turns in BiToD is 115,638, while it is 143,048 in the original English-only MultiWOZ.

\begin{figure}[!t]
    \centering
    \includegraphics[width=0.78\linewidth]{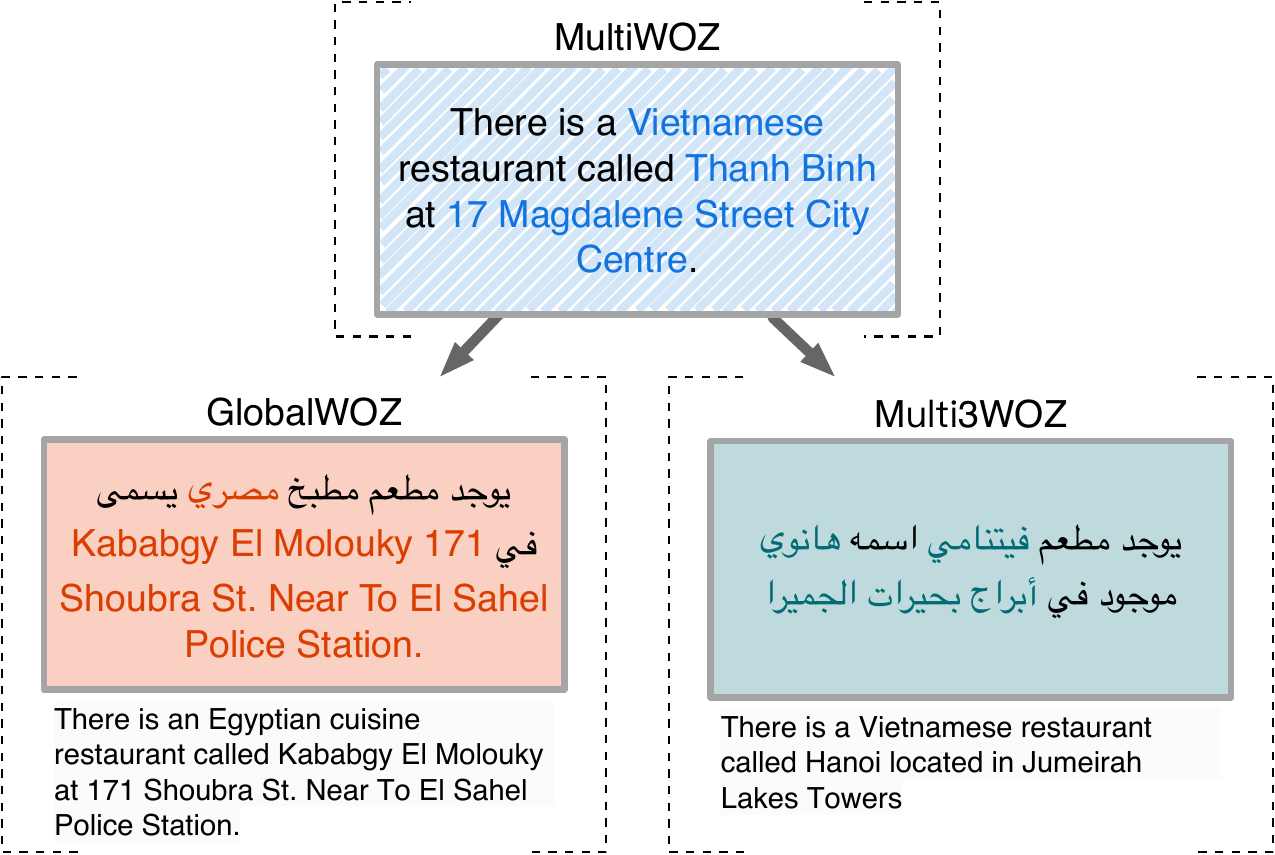}
    \caption{An example of dialog turns from culturally adapted GlobalWOZ versus \dataset, with culturally specific entities highlighted and English translations provided below each text box. In general, due to its design, a proportion of GlobalWOZ dialogs exhibit inconsistent similar code-switched and script-switched utterances (e.g., also with phone and reference numbers); GlobalWOZ comes with other design-triggered dialog-level inconsistencies, not shown for brevity.}
    \label{fig:globalwozex}
\end{figure}

\vspace{0.5mm}
\noindent \textbf{P4. (Improved) Cultural Adaptation.} 
A large number of datasets for multilingual NLP ignores the fact that the data should also be adapted to the target cultures and concepts~\cite{ponti-etal-2020-xcopa,Herscovich:2022acl}. Besides (i) propagating the source language bias towards possible \textit{conversational concept}s (e.g., the US-tied concept of \textit{tailgating} or conversations about \textit{baseball}) \cite{ponti-etal-2020-xcopa}, the lack of the so-called \textit{cultural adaptation} also (ii) creates peculiar or more unlikely \textit{conversational contexts} (e.g., a user speaking to a Turkish \tod system about restaurants in Cambridge) \cite{ding-etal-2022-globalwoz}, or (iii) even ignores specificities of a particular culture (e.g., postcodes are not used in Arabic-speaking countries). The only two datasets that try to incorporate the notion of cultural adaptation into their design are BiToD and GlobalWOZ (see Table~\ref{tab:summary_dataset}). However, BiToD's adaptation is based on a very specific bilingual region of the world (Hong Kong), while GlobalWOZ's automatic cultural adaptation approach results in a large number of incoherent dialogs and annotation errors, e.g., see Figure~\ref{fig:globalwozex}. We thus adopt a new and improved cultural adaptation approach that ensures high-quality, coherent and multi-parallel dialogs across languages while respecting the underlying cultural traits, see \S\ref{s:dataset} later.

\vspace{0.5mm}
\noindent \textbf{P5. Dialog Coherence and `Multi-Parallelism'.}
Finally, due to their design properties and oversimplifying assumptions, some datasets break coherence and multi-parallelism of dialogs. GlobalWOZ, while performing a form of cultural adaptation, (i) creates erroneous slot value annotations that are inconsistent with the dialog ontology and database in the particular language, and (ii) even induces inconsistent annotations within an individual dialog. Another problem with GlobalWOZ is that the authors select a subset of 500 test set dialogs for human PEMT work based on a simple heuristic: they opt for dialogs for which the sum of corpus-level
frequencies of their constitutive 4-grams, normalized by dialog length, is the largest. This selection, not motivated in the original paper and performed independently for each language, entails that different portions of the original English MultiWOZ are included into the final language-specific test sets. This design choice, besides (i) artificially decreasing linguistic diversity of dialogs chosen for the test set in 
each language,\footnote{The selection heuristic favors dialogs that contain
the same most frequent 4-grams globally.} also (ii) breaks the desired multi-parallel nature of the test set. As a consequence, GlobalWOZ overestimates downstream \tod performance for target languages, and cannot be used for any direct comparison of \tod task performance across different languages since test sets per language contain different dialogs, as also pointed out by \citet{hung-etal-2022-multi2woz}.

\dataset is the only dataset which performs cultural adaptation and avoids confouding factors such as GlobalWOZ's selection heuristics, while maintaining the desired properties of dialog coherence and multi-parallelism.

\begin{figure*}[t!]
    \centering
    \includegraphics[width=0.85\linewidth]{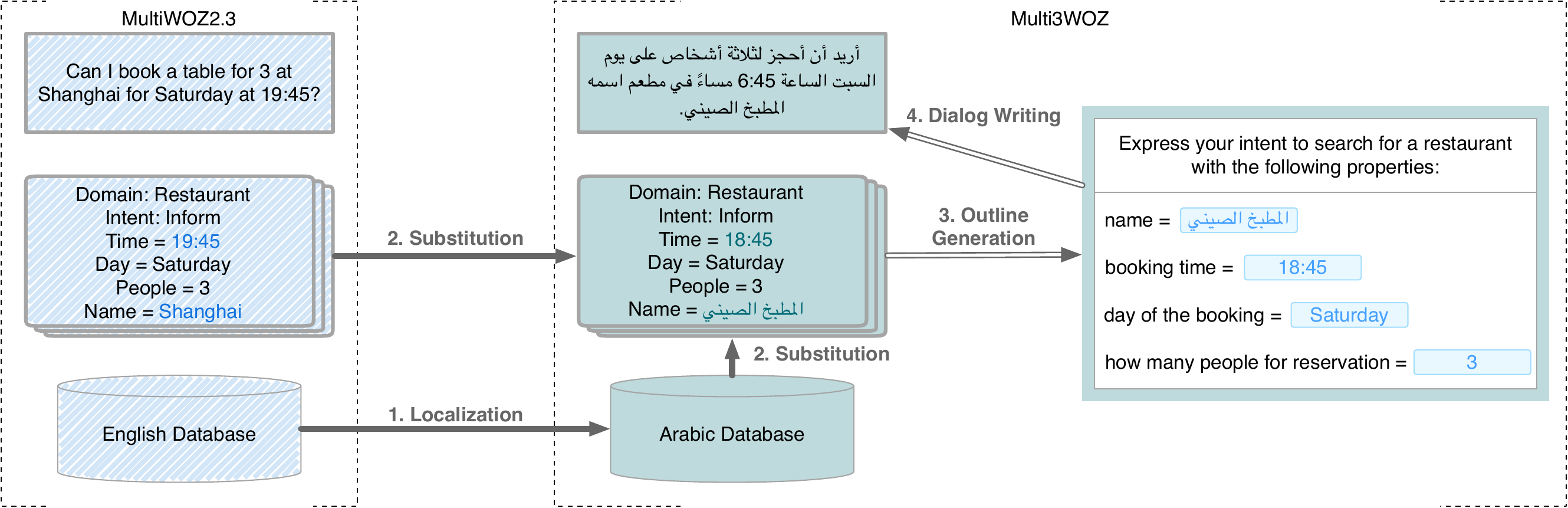}
    \caption{Overview of the full data collection pipeline for \textcolor{wozgreen}{\dataset}. It is derived from the \textcolor{wozblue}{MultiWOZ} dataset v2.3, with two phases: (i) \textit{cultural adaptation} and (ii) \textit{outline-based generation}. Cultural adaptation $\rightarrow$ spans two subtasks \textit{localization} and \textit{value substitution}, and it adapts dialogs and contextualizes them to the actual existing entities from the cultures in which the target languages are spoken. Outline-based generation $\Rightarrow$ is a bottom-up dialog collection method to collect language-specific and linguistically adapted surface forms from the target language native speakers based on language-agnostic abstract dialog schemata.
    In both datasets, each utterance is annotated with task-specific meaning representations. In the above figure, a rectangle \raisebox{-0.1\totalheight}{\includegraphics[scale=0.1]{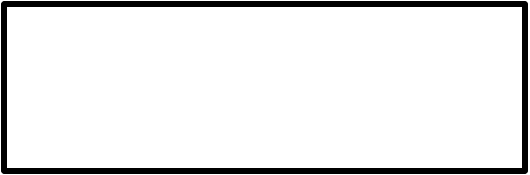}} denotes an utterance and stacked rectangles \raisebox{-0.3\totalheight}{\includegraphics[scale=0.1]{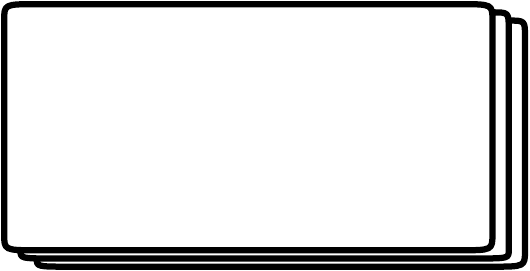}} denote its corresponding dialog act. Further, each dialog is conditioned on a culture-adapted ontology database \raisebox{-0.2\totalheight}{\includegraphics[scale=0.1]{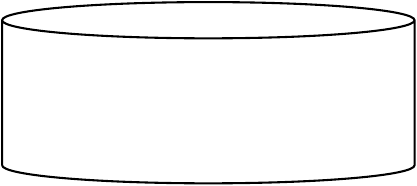}} as an extra-linguistic context, and it must be coherent with the database content.}
    \label{fig:fig1}
\end{figure*}

\section{\dataset}
\label{s:dataset}

\dataset comprises linguistically and culturally adapted task-oriented dialogs in four languages:  Arabic (\ara; Afro-Asiatic), English (\eng; Indo-European), French (\fra; Indo-European), and Turkish (\tur; Turkic). A total of 27,480 (3$\times$9,160) dialogs is collected for \ara, \fra, \tur, while the dataset also includes a subset of 9,160 normalized and corrected MultiWOZ v2.3 dialogs.\footnote{We select 9,160 out of MultiWOZ's full set of 10,438 dialogs by filtering out erroneous dialogs identified during the normalization and cultural adaptation process; problematic dialogs were also recorded by our annotators during the dialog generation and quality control phases (see later in \S\ref{s:dataset}).}

In what follows, we describe its creation, as depicted in Figure~\ref{fig:fig1}. Our approach involves three key steps:
\begin{enumerate*}[label=(\roman*)]
\item \textit{normalizing annotations} from the original MultiWOZ v2.3 with canonical values.
\item \textit{cultural adaptation} by contextualizing dialogs to entities from the relevant cultures.
\item \textit{collecting linguistically adapted dialogs} from target language native speakers using a bottom-up outlined-based method.
\end{enumerate*}

\rparagraph{Preliminaries and Notation}
In \tod, the domains of a dataset (e.g., MultiWOZ) and the systems built upon it are typically defined by an \textit{ontology}, which provides a structured representation of an underlying \textit{database}. The ontology specifies slots that encompass all entity attributes and their corresponding values~\cite{Budzianowski:2018multiwoz}.
\dataset is designed to be fully compatible with the original English MultiWOZ's ontology and data format, but now with culturally adapted database entries (see Figure~\ref{fig:fig1}).

\dataset $\mathbbm{D}$ contains four multi-parallel sets of dialogs, namely $\mathbbm{D}^{\ara}$, $\mathbbm{D}^{\eng}$, $\mathbbm{D}^{\fra}$, and $\mathbbm{D}^{\tur}$, along with their corresponding \textit{cultural-specific databases} denoted as $\mathbbm{E}^{\ara}$, $\mathbbm{E}^{\eng}$, $\mathbbm{E}^{\fra}$, and $\mathbbm{E}^{\tur}$.\footnote{In order to simplify our notation, we represent a backend database as a set of data entries, where each entry corresponds to a real-world entity within the target culture.} Each database entry, $\mathcal{E} \in \mathbbm{E}$, contains a set of slot-value pairs, such that $\mathcal{E} = \{({s}_{1}, {v}_{1}), ({s}_{2}, {v}_{2}),\cdots,({s}_{n}, {v}_{n})\}$.\footnote{We denote each attribute of an entity as a slot and consider the domain of an entity as an inherent attribute. 
For example, \textit{\{(domain, police), (name, parkside police station), (address, Parkside, Cambridge), (phone, 01223358966), (postcode, cb11jg)\}} is a database entry in $\mathbbm{E}^{\eng}$.}
Each dialog in the dataset is represented as a list of natural language utterances, with alternating turns between the user and system initiated by the user. Each turn is annotated with its corresponding sentence-level meaning representation.
Namely, for $\mathcal{D} \in \mathbbm{D}$, $\mathcal{D} = [(\mathbf{u}_{1}, \mathbf{a}_1),  \cdots, (\mathbf{u}_{j}, \mathbf{a}_j)]$, where $\mathbf{u}$ is a surface form (user or system) utterance; $\mathbf{a}$ is a dialog act representation; $j$ is the length of the dialog $\mathcal{D}$.

A dialog act $\mathbf{a}$ is then defined as a set of tuples $\mathbf{a} = \{(d_1, i_1, s_1, v_1), \cdots, (d_k, i_k, s_k, v_k)\}$, where each tuple consists of domain $d$, intent $i$, slot $s$, and slot value $v$.

\rparagraph{Slot-Value Normalization}
In the English MultiWOZ dataset, slot values are annotated as text spans within the corresponding utterances. This annotation scheme allows for more flexible and natural language expressions of the canonical value $v_{\text{truth}}$ described in the ontology and database (e.g., \textit{13:00}), resulting in various surface forms ${v^{(1)}, \cdots, v^{(l)}}$ (e.g., \textit{1 pm}, \textit{1:00 pm}, \textit{one}). However, this flexibility can create a discrepancy between the expected canonical value required by the backend API and the predicted value by the model.\footnote{The query sent to the backend API is formulated using a formal language that lacks the flexibility of natural language. This issue can significantly affect the performance of extractive models, such as extractive DST models~\cite{heck-etal-2020-trippy, zhou-etal-2023-xqa}.}

Moreover, the absence of a 1-to-1 mapping between the canonical value in the database and the annotations in MultiWOZ, coupled with erroneous or misspelled entries, hinders the consistent and systematic adaptation of culture-dependent entities to the target language.
To address this, we \textit{manually} created a normalization dictionary and assigned canonical values to all slot values across the English MultiWOZ dataset. For example, we created a normalization dictionary for the \textit{restaurant-name} slot, mapping 544 distinct surface forms to 110 canonical names. These canonical names correspond exactly to the entities in the English \textit{restaurants} domain's database, enabling a one-to-one mapping between the entities described in dialogs and those in the database. Besides facilitating cultural adaptation through the creation of surface form agnostic outlines, we believe that this time-consuming yet crucial normalization process will also enable consistent evaluations of models built on \dataset. Henceforth, any mention of a slot value $v$ assumes that it is in its canonical form.\footnote{The introduction of slot values in canonical forms offers supplementary information to the original MultiWOZ annotation. The original format can be automatically derived, enabling backward compatibility with previous models.}

\rparagraph{Cultural Adaptation}
While English MultiWOZ contains only dialogs describing entities in the Cambridge (UK) area, \dataset expands the scope to three additional languages targeting three cities where the target languages are considered native: Dubai for Arabic, Paris for French, and Ankara for Turkish.\footnote{We fully acknowledge that here we use the term `culture' (imprecisely) as a proxy for the limited set of properties, customs, and entities to be expected or common at the target location. We also acknowledge that language-culture mappings are typically many-to-many, with the possibility of multiple languages being native to the same culture, and one language spreading over more than one culture or subculture \cite{Herscovich:2022acl}. Our (simplified) choice is primarily driven by pragmatic considerations and feasibility requirements.} 
To ensure that our dataset respects and reflects the cultural traits pertaining to each target city and language, we propose a systematic approach for cultural adaptation, which ensures dialog coherence and multi-parallelism across all languages, and includes the following steps: 
\begin{enumerate*}[]
\item \textit{slot-value localization/redistribution} with cultural awareness,
\item \textit{controlled entity replacement} with one-to-one entity mappings,
\item \textit{slot-value randomization} to avoid verbatim memorization.
\end{enumerate*}

We perform \textit{slot-value redistribution} to adjust the original slot and value to align with the target `culture'. These modifications are based on the feedback from native speakers of the target language with expertise in the corresponding cultural context. To better fit the target culture, we remove \eng-specific slots and values that are irrelevant to the culture. For example, we obliterate the \textit{postcode} slot in the Arabic dataset $\mathbbm{D}^{\ara}$ due to its limited relevance in the associated culture.\footnote{We also consider religious factors: e.g., to respect local culture, we replace the `gastropub' \textit{restaurant type} with the value `Arab', or `nightclub' with `waterpark' for the \textit{attractions} slot. Moreover, we address the issue of unbalanced entity distribution in the original MultiWOZ, which is heavily skewed towards Cambridge (UK) and contains a disproportionate number of mentions of `colleges' and `guest houses'.
To mitigate this bias, we swap certain types of entities; e.g., we exchange the very specific term \textit{`college'} with \textit{`architecture'} and \textit{`guest house'} with \textit{`hotel'} to offer a better localization of the entity distribution for the target location.}

The main objective of our proposed cultural adaptation method is to perform \textit{controlled entity replacement} using a 1-to-1 entity mapping.
As a prerequisite, we first construct a localized database (e.g., $\mathbb{E}^{\ara}$ for Arabic) for each target language. This database aims to reflect real-world entities and properties, and has been constructed by human participants in our project, native speakers of the target languages, who referred to a variety of public knowledge sources on the Internet, including the Google Places API and TripAdvisor API.\footnote{However, we note that, for database completeness, a portion of the entity information has been synthetically generated due to missing information on the Web, e.g., when a restaurant does not provide a phone number on its website.}

In order to construct such a 1-to-1 mapping, an English entity $\mathcal{E}^{\eng}$ and a target entity (e.g., $\mathcal{E}^{\ara}$) can be mapped to each other only if all categorical slot values attributed to each entity are identical.\footnote{A categorical slot is defined by the ontology such that the possible values for this slot are a closed set. For example, the slot `price range' can only have the values of `cheap', `moderate', and `expensive'. In contrast, the value for a \textit{hotel name} is an open set and not categorical, as it can be any string.}
Namely, the following condition holds: $\forall ({s}^{\eng}, {v}^{\eng}) \in \mathcal{E}^{\eng}, \exists ({s}^{\ara}, {v}^{\ara}) \in \mathcal{E}^{\ara} : {v}^{\eng} = {v}^{\ara} \; \text{if} \; \texttt{is\_categorical}({s}^{\eng})$.
This strategy guarantees a consistent distribution of entities with respect to each categorical property as MultiWOZ. It further facilitates the coherent and multi-parallel creation of dialogs, particularly when the user requests a certain property of a desired entity along the progress of dialogs (e.g. `an \textit{expensive} restaurant'). This stands in contrast to the random sampling cultural adaptation solution of GlobalWOZ, which results in frequently mismatched entities being returned in response to the user request, and often results in dialog incoherence.

The original MultiWOZ contains a substantial number of randomized slot values, such as \textit{time}, \textit{reference}, and \textit{taxi-phone}. To prevent verbatim memorization and undesired data artefacts, we perform \textit{slot-value randomization} independently in each target dialog subset in \dataset.
For \textit{time}-related slot values in \dataset, we apply the randomization by adding a 1-hour random offset drawn from a uniform distribution [-1, 1] to the original value, as also illustrated in Figure~\ref{fig:fig1}.
We ensure that all \textit{time} relevant slots (e.g. \textit{leaving time} and \textit{arriving time}) in a dialog are equivalently shifted by the same randomized offset. For \textit{reference} numbers, we employ the 1-to-1 randomly generated reference mapping.
Regarding \textit{taxi-phone} values, we first adhere to the target culture's specific phone pattern followed by a 1-to-1 randomly generated phone mapping. 
In general, this procedure mitigates the risk of exploiting annotation artifacts and consequent overfitting when conducting cross-lingual transfer learning experiments.

\rparagraph{Outline-Based Dialog Generation}
By adopting the outline-based dialog generation process we simultaneously enable cultural adaptation while eliminating the impact of syntactic and lexical grounding in the source language (i.e., the so-called ``translation artifacts''), while keeping the annotation protocol feasible \cite{Majewska:2023cod}.
The outline-based method can be decomposed into two steps: \textit{outline creation} (i.e., creating dialog schemata) and \textit{dialog writing} (i.e., creating the actual surface realizations, utterances, from the dialog schemata).

Following \citet{Majewska:2023cod}, \textit{outline creation} involves creating minimal but comprehensive instructions for the so-called \textit{dialog creators} (termed DCs henceforth) to generate dialogs that fully convey specific intents and slots while avoiding the imposition of predefined syntactic structures or linguistic expressions. 
As depicted in Figure~\ref{fig:fig1}, we convert a culturally adapted (termed \textit{CA-ed} henceforth) %
dialog act (e.g., using \ara as an example language, $\mathbf{a}^{\ara}$) into a human-interpretable outline based on a set of manually defined templates, where different sets of templates are used for the user and system utterances. Given a tuple $(d, i, s, v^{\ara}) \in \mathbf{a}^{\ara}$, we transform a domain-intent pair $d$-$i$ into a natural language instruction, e.g., \texttt{Restaurant-Inform} $\Rightarrow$ \textit{``Express your intent to search for a restaurant with the following properties:''}. In addition, the slot $s$ is mapped to a predefined natural language description, and it is presented along with the CA-ed slot value $v^{\ara}$ (e.g., \textit{booking time = 18:45}).
As illustrated in Figure~\ref{fig:fig1}, in cases where there are multiple tuples with the same pair $d$-$i$, we group them together and present within a \textit{``card''}.
We note that a target language utterance (e.g., $\mathbf{u}^{\ara}$) can be constructed based on multiple cards, with each card corresponding to a unique domain-intent pair $d$-$i$.\footnote{\textit{Restaurant-Inform} is the domain-intent pair for the utterance \textit{There will be 5 of us and 19:45 would
be great.}}
Besides, each card may contain multiple slot-value pairs, where each slot value is shown as a CA-ed value (e.g., $v^{\ara}$).
To take full advantage of our outline-based framework, we have developed a Web-based annotation toolkit along with detailed annotation guidelines; the latter is made publicly available.

\textit{Dialog writing} is then carried out by bilingual speakers as DCs. They are (i) native in the target language and (ii) fluent in English: following the results from our pilots, we opted for keeping the English templates as it facilitated the quality control of templates and cards while it did not have any detrimental effect on the quality of finally generated target language dialogs. The DCs were instructed to write natural-sounding exchanges in their native language between a hypothetical user and an assistant, based on the outlines derived from the CA-ed dialog act (e.g. $\mathbf{a}^{\ara}$) and a set of user goals that the hypothetical user wants to achieve (e.g., \textit{You are looking for a place to stay.}). 
For each utterance $\mathbf{u}$ from the source \eng dataset, the tasks of the DCs were then as follows: 
\textbf{1)} writing a native dialog utterance from the card(s) that covers all the slot values from the cards;
\textbf{2)} annotating character-level span indices for each slot value $v^{\ara}$;
\textbf{3)} indicating with a binary flag for each domain-intent pair $d$-$i$ whether this dialog act retains coherence of the full dialog, this way also signaling and capturing errors still present in the English MultiWOZ v2.3 dataset.

\begin{figure*}[t!]
    \centering
    \includegraphics[width=0.95\linewidth]{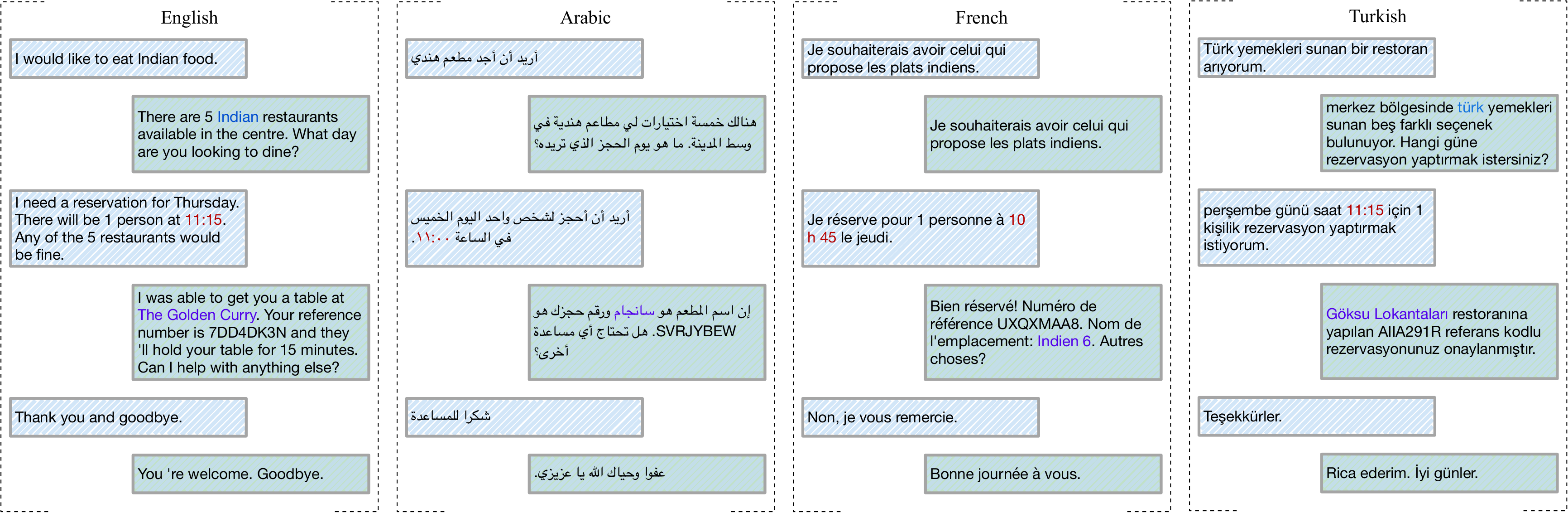}
    \caption{An example set of parallel dialogs in four languages: English, Arabic, French, and Turkish, extracted from the \dataset dataset. The dialogs illustrate different aspects of cultural adaptation, including \textcolor{wozverybleu}{slot-value redistribution}, \textcolor{wozveryred}{slot-value randomization}, and \textcolor{wozverypurple}{controlled entity replacement}, which are highlighted with distinct colors. Due to space limitations, we only show a set of single-domain short dialogs. However, it is important to note that the \dataset dataset contains multi-domain dialogs with diverse dialog patterns and linguistic constructions. The dialog ID for this specific example is SSNG0101.}
    \label{fig:examples}
\end{figure*}

\rparagraph{Duration, Cost, Dialog Creators, Quality Control}
The logistically and technically complex data collection process spanned 14 months, starting in January 2022. The full cost of data collection was $\sim$\$64,500, equally distributed across the three target languages.  
The recruited DCs are (i) professional translators and (ii) college students, recruited via the ProZ platform (\url{www.proz.com}) or from universities worldwide. A total of 133 native Arabic speakers, 112 native French speakers, and 75 native Turkish speakers contributed to the dataset.

We applied a number of quality control mechanisms throughout the data collection process. First, to ensure that the DCs have fully understood the instructions and all (sub)tasks, they were required to complete a qualification round before creating any actually deployed data. Second, our annotation platform features a real-time automatic check for all submissions, providing feedback and highlighting issues for the collected dialogs. Finally, we also ran two rounds of \textit{post-collection dialog editing}: we invited a carefully selected small group of dialog creators, who had consistently produced exceptional high-quality dialogs, to review and, if necessary, edit all the dialogs in the validation and test sets of all three target languages.

\rparagraph{Ethical and Responsible Data Creation and Use}
Following the principles from~\citet{rogers-etal-2021-just-think}, the project has placed a high priority on ethical and responsible data creation and use. It underwent the full Ethics Approval process at University of Cambridge, and we describe other ethics-related aspects here. %

\rrparagraph{Terms of Use}
\dataset is released under the same MIT License as the original MultiWOZ.

\rrparagraph{Privacy}
To comply with the EU General Data Protection Regulation (GDPR), we have acted as a data controller and collected the minimum of personal data required for this project. 
All participants provided informed consent by signing a \textit{Participant Consent Form} before any data collection occurred. %
To adhere to the principle of data minimization, we collected only the participants' email addresses as individually identifiable information for the sole purpose of processing payments.
Our dataset consists solely of hypothetical dialogs in which the domains and content have been restricted and predefined, minimizing the risk of personal data being present in \dataset.

\rrparagraph{Compensation}
The DCs were compensated based on the number of dialogs they contributed to the dataset, with a payment rate of approximately \$12/h. As stated in our consent form, they were able to withdraw from the study at any time. %

\rparagraph{Data Structure and Statistics}
Figure~\ref{fig:examples} presents an example of multi-parallel dialogs from \dataset. All dialogs in \dataset consist of parallel surface form utterances in multiple languages and retain the same annotations as the original MultiWOZ. Precisely, each dialog  $\mathcal{D}$ is annotated with a CA-ed user goal, as well as for each utterance $\mathbf{u}$ in the dialog: a CA-ed dialog act, a CA-ed dialog state. In addition, \dataset offers (i) annotations for character-level textual spans for all the slot values in the dialog act to steer span extraction-based solutions to slot labeling \cite{joshi-etal-2020-spanbert}, and (ii) a binary coherence indicator. The dataset is released in three standard formats: (i) \texttt{json} files following the structure of MultiWOZ \cite{Budzianowski:2018multiwoz}; (ii) a format compatible with the Huggingface repository \cite{wolf-etal-2020-transformers,lhoest-etal-2021-datasets}; (iii) ConvLab-3-compatible format \cite{convlab-3}.

\begin{figure}[t!]
    \centering
    \includegraphics[width=0.9\linewidth]{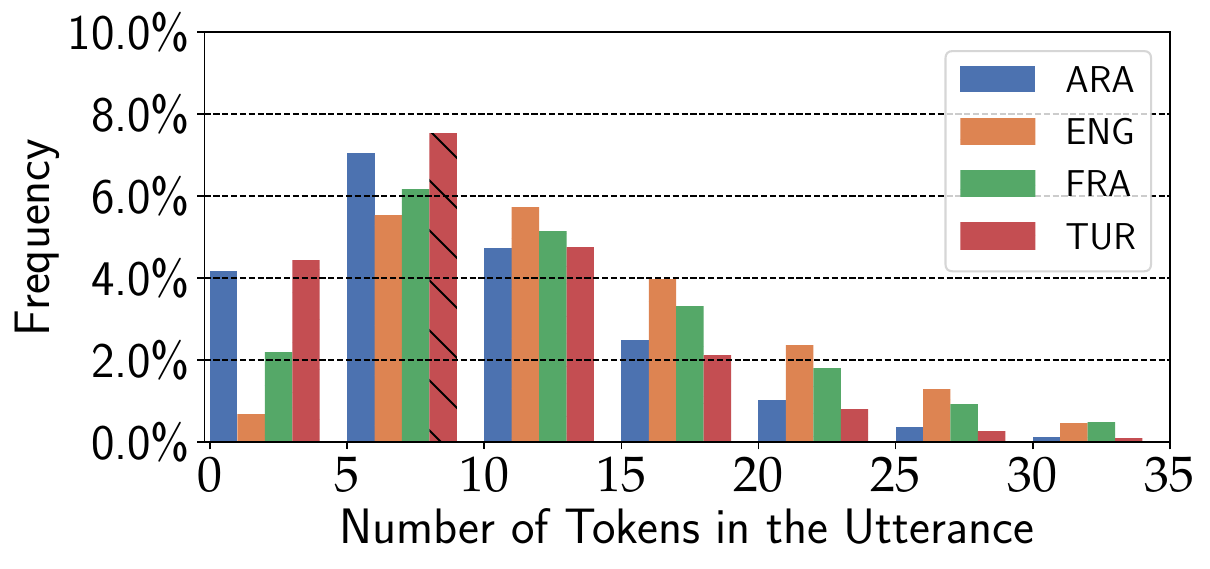}
    \caption{Utterance length in \dataset.}
    \vspace{-1mm}
    \label{fig:fig2}
    \vspace{-1.5mm}
\end{figure}

\dataset's language-independent features, e.g. the frequency of dialog acts and average dialog length, closely resemble those of the original MultiWOZ; we thus focus on the statistics pertaining to language and cultural adaptation. %
Figure~\ref{fig:fig2} presents the distribution of the number of tokens per turn, with white spaces as the token delimiter. Note that each language exhibits variance in its morphosyntactic properties (e.g., Turkish is an agglutinative language), which naturally impacts the expected utterance length. Further, we find that 13.3\% of the slot values in the dialog acts are normalized with canonical values, while 38.7\% of the dialog acts' slot values are provided with CA-ed values. The type-to-token ratio (TTR) varies across languages, with English having a lower TTR value (0.010) compared to Arabic (0.032), French (0.023), and Turkish (0.035). In comparison to the GlobalWOZ dataset, which is an MT-based dataset without CA, our dataset (\dataset) achieves an increased TTR for Arabic ($\uparrow$ 0.013), French ($\uparrow$ 0.006), and Turkish ($\uparrow$ 0.014).\footnote{For this comparison, we utilize the ``F\&E'' proportion of the GlobalWOZ dataset. In this dataset, English utterances are translated into the target language using Google Translate, while preserving the slot values associated with English entities. The calculation of the TTR is limited to the dialogs that are included in both the GlobalWOZ dataset and our dataset.} This outcome highlights that \dataset's bottom-up design sparked higher semantic variability and naturalness in the target languages~\cite{Majewska:2023cod}.
We further highlight the higher semantic diversity of utterances in \dataset in comparison to PEMT-based methods such as the one used by Multi$^2$WOZ. We select a subset of 1,586 Arabic dialogs of flows shared between the two datasets and calculate the average pairwise cosine similarity between utterances in each data subset and their corresponding utterances in the English MultiWOZ, relying on LaBSE \cite{labse} as a state-of-the-art multilingual sentence encoder.
The scores of 0.54 (\dataset) and 0.91 (Multi$^2$WOZ) suggest the higher semantic variability created through the outline-based approach with cultural adaptation.

\section{\dataset as a \tod Benchmark}
\dataset establishes a multilingual and cross-lingual benchmark for \tod systems and their sub-modules. We now present a first `benchmarking study' on the dataset, evaluating representative models for NLU, DST, NLG, and E2E tasks in \tod, merely scratching the surface of possible experimental work now enabled by \dataset.

\begin{table*}[!t]
\centering
\def\arraystretch{0.67}
{\scriptsize
\resizebox{0.77\textwidth}{!}{%
\begin{tabular}{@{}lccccccccccc@{}}
\toprule
                           &  & \multicolumn{2}{c}{Intent Detection} &  & \multicolumn{3}{c}{Slot Labeling} &  & \multicolumn{3}{c}{Dialog State Tracking} \\ \cmidrule(lr){3-4} \cmidrule(lr){6-8} \cmidrule(l){10-12} 
\multirow{-2}{*}{Language} &  & Accuracy           & F1              &  & Precision    & Recall    & F1      &  & JGA         & Turn Acc.      & F1         \\ \midrule
\multicolumn{12}{c}{\cellcolor[HTML]{EFEFEF}\textbf{Fully Supervised (Monolingual)}}                                                                                                \\ \midrule
ENG                        &  & 92.71              & 95.77           &  & 95.92        & 94.08     & 94.99   &  & 59.90       & 97.87          & 93.67      \\
ARA                        &  & 92.20              & 94.59           &  & 48.44        & 42.47     & 45.26   &  & 47.72       & 96.85          & 89.26      \\
FRA                        &  & 88.92              & 92.93           &  & 79.57        & 77.76     & 78.65   &  & 49.77       & 97.02          & 89.93      \\
TUR                        &  & 91.50              & 94.52           &  & 87.25        & 86.72     & 86.94   &  & 53.59       & 97.28          & 91.04      \\
AVG.                       &  & 91.33              & 94.45           &  & 77.80        & 75.26     & 76.46   &  & 52.74       & 97.26          & 90.97      \\ \midrule
\multicolumn{12}{c}{\cellcolor[HTML]{EFEFEF}\textbf{Zero-Shot Cross-lingual Transfer} (from English)}                                                                               \\ \midrule
ARA                        &  & 60.28              & 71.06           &  & 17.56        & 27.74     & 21.47   &  & 1.47        & 80.73          & 5.80       \\
FRA                        &  & 72.88              & 81.71           &  & 48.53        & 60.52     & 53.86   &  & 3.66        & 85.08          & 32.83      \\
TUR                        &  & 69.52              & 79.09           &  & 48.47        & 66.80     & 56.18   &  & 1.30        & 82.05          & 15.22      \\
AVG.                       &  & 67.56              & 77.29           &  & 38.19        & 51.69     & 43.84   &  & 2.14        & 82.62          & 17.95      \\

\bottomrule

\end{tabular}
}
}%
\caption{Fully supervised and zero-shot cross-lingual transfer from English ($\mathbbm{D}^{\eng}$ as the source) for ID, SL, and DST tasks on \dataset. AVG. shows the mean average of the evaluation scores across all four languages. The reported scores are averaged over 3 random runs.}
\label{tab:nlu-results}
\end{table*}

\begin{table*}[t!]
\centering
\def\arraystretch{0.67}
{\scriptsize
\resizebox{\textwidth}{!}{%
\begin{tabular}{@{}llccccccccccc@{}}
\toprule
\multirow{2}{*}{Language} &  & \multicolumn{3}{c}{Surface Realization} &  & \multicolumn{3}{c}{Language Modeling} &  & \multicolumn{3}{c}{Language Modeling with Oracle} \\ \cmidrule(lr){3-5} \cmidrule(lr){7-9} \cmidrule(l){11-13} 
                          &  & BLEU        & ROUGE       & METEOR      &  & BLEU       & ROUGE       & METEOR      &  & BLEU            & ROUGE          & METEOR          \\ \midrule
ENG                       &  & 20.67       & 47.76       & 44.16       &  & 8.66       & 27.95       & 25.18       &  & 21.20           & 48.52          & 44.31           \\
ARA                       &  & 9.57        & 14.04       & 21.92       &  & 7.22       & 20.77       & 18.11       &  & 17.56           & 15.99          & 35.22           \\
FRA                       &  & 9.96        & 35.31       & 29.17       &  & 6.19       & 24.47       & 19.78       &  & 13.61           & 40.69          & 34.87           \\
TUR                       &  & 13.59       & 39.29       & 33.99       &  & 9.87       & 30.07       & 26.84       &  & 24.23           & 53.76          & 48.49           \\
AVG.                      &  & 13.45       & 34.10       & 32.31       &  & 7.98       & 21.14       & 22.48       &  & 19.15           & 39.74          & 40.72           \\ \bottomrule
\end{tabular}%
}
}%
\caption{Fully supervised NLG performance for mT5$_\textit{small}$. AVG. shows the mean average of the evaluation scores across all four languages. The reported scores are averaged over 3 random runs.}
\label{tab:nlg_results}
\end{table*}

\rparagraph{Natural Language Understanding}
NLU is typically decomposed into two established tasks: intent detection (ID) and slot labeling (SL). ID can be cast as a multi-class classification task that identifies the presence of a domain-intent pair $d$-$i$ (e.g., \texttt{Restaurant-Inform}) from the user's utterance, where the set of intents is predefined in the ontology. SL is a sequence tagging task that identifies the presence of a value $v$ and its corresponding slot $s$ within the utterance.

We evaluate ID and SL methods backed by XLM-R$_\textit{base}$ \citep{conneau-etal-2020-unsupervised}.
Precisely, at each dialog turn $t$, the model encodes the concatenation of the previous two utterances ($\mathbf{u}_{t-2}$ and $\mathbf{u}_{t-1}$) along with the current utterance ($\mathbf{u}_{t}$) to provide embedding vectors at both the sequence and token levels.
To implement the intent detector, for each domain-intent pair $d$-$i$ defined by the ontology, the representation of the ``\texttt{<s>}'' token is subsequently projected down to two logits and passed through a Sigmoid layer to form a Bernoulli distribution indicating if $d$-$i$ appears in the $\mathbf{u}_{t}$. Performance is evaluated by measuring its accuracy in identifying the exact presence of all domain-intent pairs in a dialog act, as well as its F1 score. For SL, we adopt the widely-used BIO labeling scheme to annotate each token in the user's utterance.\footnote{Specifically, each token is labeled with either B-$d$-$i$-$s$ (e.g., \textit{B-Restaurant-Inform-Food}), denoting the beginning of a slot-value pair with the corresponding slot name, I-$d$-$i$-$s$ indicating it is inside the slot-value, or O indicating that the token is not associated with any slot-value pair.}
\footnote{We conducted all NLU experiments on a single RTX 24 GiB GPU with a batch size of 64 and a learning rate of $2e-5$. We trained each model for 10 epochs and selected the model with the best F1 score on the validation set as the final model.}

In Table~\ref{tab:nlu-results}, we observe that the fully supervised ID model achieves similarly high accuracy across all languages, and we also observe a large cross-lingual transfer gap~\cite{Hu:2020xtreme} for both tasks. Further, there is a substantial decrease in performance for Arabic SL.
Note that in \dataset the slot-value spans are annotated at the character level, and we only consider a span to be correctly identified if there is an exact match. At the same time, \citet{rust-etal-2021-good} observed that the sub-optimal performance of the tokenizers for the multilingual models may yield degraded downstream performance. To investigate the limitations of tokenization, we then aligned the slot boundaries with the token boundaries. Specifically, we defined the slot span as the minimal token span that covered the entire slot in the utterance. With this approach, the identical model achieved F1 of 78.44 ($\uparrow$30.00) for Arabic SL, confirming that the suboptimal XLM-R's tokenization was the primary contributor to the original performance degradation in Arabic.

\rparagraph{Dialog State Tracking}
For DST, we follow the standard MultiWOZ preprocessing and evaluation setups \citep{wu-etal-2019-transferable}, excluding the `hospital' and `police' domains due to the absence of test dialogs in these domains. We report the Joint Goal Accuracy (JGA), Turn Accuracy, and Joint F1.

We adapt T5DST \citep{lin-etal-2021-leveraging} as a strong baseline that reformulates the DST as a QA task with slot descriptions. The DST model is back-boned with mT5$_\textit{small}$ \citep{xue-etal-2021-mt5} (as very similar scores were obtained with mT5$_\textit{base}$).
Regarding the model and training details, readers are referred to the original work \citep{lin-etal-2021-leveraging}.\footnote{The experiments were run on a single RTX 24 GiB GPU, a batch size of 4 and a learning rate of $1e-4$; 5 epochs.}

Fully supervised DST scores provide a strong benchmark with the multilingual T5DST model over all languages in \dataset. We observe the highest performance in English ($59.9\%$ JGA), followed by Turkish, French, and Arabic, indicating the levels of difficulty of DST for each language. Table~\ref{tab:nlu-results} presents the zero-shot cross-lingual transfer-from-English results, revealing poor transferability of the DST models across languages (all below $4\%$ JGA). This indicates the limitations of current multilingual models in zero-shot setups and the challenge of transfer learning for culturally adapted dialogs in \dataset.

\rparagraph{Natural Language Generation}
We approach the NLG task as a sequence-to-sequence problem, again supported by mT5$_\textit{small}$. Specifically, at each dialog turn $t$, the model takes the input of its dialog context, and generates a system response $\mathbf{u}_{t}$. Traditionally, NLG in \tod systems is defined as the task of converting a dialog act into a natural language utterance~\cite{williams2007partially}. In our study, we evaluate NLG performance in both a traditional setup, where the goal is to realize the {surface form} of the dialog act, and an end-to-end LM setup, where we model response generation as a transduction problem from the dialog history to a natural response. Third, we consider the setup where both the dialog history and the `oracle' dialog act are available, serving as a performance upper bound. %
For the \textit{surface realization} setup, we convert the dialog act $\mathbf{a}_{t}$ into a flattened string format (e.g., \textit{[inform][restaurant]([price range][expensive],[area][center]}) to serve as the input. For the \textit{language modeling} setup, the model generates a response $\mathbf{u}_{t}$ solely based on the preceding dialog history $\mathbf{u}_{t-2}$ and $\mathbf{u}_{t-1}$. In this setup, the generation model does not have any knowledge about the system's ontology and database. In the \textit{language modeling with oracle} setup, the model takes the concatenation of the two preceding utterances $\mathbf{u}_{t-2}$ and $\mathbf{u}_{t-1}$, as well as $\mathbf{a}_{t}$ as input.

Following MultiWOZ, we evaluate with the corpus BLEU score~\cite{papineni2002bleu}; we evaluate lexicalized utterances without performing delexicalization.
We also report ROUGE-L~\cite{lin-2004-rouge} and METEOR~\cite{banerjee2005meteor}.%
\footnote{All NLG experiments were run on a single A100 80 GiB GPU; batch size of 32, a learning rate of $1e-3$; 10 epochs.}

The results are summarized in Table~\ref{tab:nlg_results}. We observe that the performance of English is significantly higher than other languages in the first setup. This disparity can be attributed to the fact that dialog acts are considered a formal language for the system to process internally and, except for culturally adapted values, they are provided in English. Therefore, it is more challenging for a model to learn how to generate natural language utterances in other languages. Furthermore, by incorporating the dialog history and the oracle dialog act, the performance of all three languages improved significantly, indicating that modeling the dialog history contributes to more coherent responses. Lastly, in the absence of database information, the performance for all languages is considerably low. This highlights the challenge of modeling \tod, and underlines the necessity of incorporating databases into the \tod models in future work.

\rparagraph{End-to-End Modeling}
Finally, E2E modeling performance serves as an even more comprehensive, challenging and arguably more important indicator for assessing the progress of \tod research, garnering intensified research attention~\cite[\textit{inter alia}]{10.5555/3495724.3497418, lin-etal-2020-mintl, peng-etal-2021-soloist, su-etal-2022-multi, wu2023using}. Developing an E2E system offers several advantages 
over focusing on individual sub-components like NLU modules or dialog state trackers. The E2E approach achieves increased applicability, enabling the development of practical real-world applications.
Moreover, it reduces vulnerability to error propagation across sub-components and offers a simpler system design compared to the traditional pipelined approaches.

To the best of our knowledge, no previous publicly available implementation of a multilingual E2E \tod system exists that would be compatible with the MultiWOZ dataset and its derivatives. Other available multilingual \tod benchmarks either lack E2E results~\cite{hung-etal-2022-multi2woz, ding-etal-2022-globalwoz}, or do not release their implementation~\cite{Zuo:2021allwoz}. The only exception is BiToD~\cite{Lin:2021bitod}; however, the BiToD dataset and the associated system use a different annotation schema, which is incompatible with MultiWOZ. Therefore, we present the first publicly available implementation of a multilingual E2E system compatible with the MultiWOZ-related datasets. We release this implementation as a baseline for further research and experimentation on \dataset.

Our system is composed of three key components: a Dialog State Tracking (DST) model, a Database (DB) Interface component, and a Response Generation (RG) model. First, the DST model is a sequence-to-sequence model, which takes the concatenated lexicalized form of all the historical utterances as input and generates a linearized dialog state (e.g., \textit{hotel price range = cheap ; type = hotel}).
Then, the DB Interface transforms the predicted dialog state into an SQL query. This query is executed, resulting in a list of entities that satisfy the specified constraints, which are then returned to the system. Finally, the RG model, also implemented as a seq2seq model, takes as input the concatenation of historical utterances, predicted dialog state, and a database summary that indicates the number of entities returned for each active domain (e.g., \textit{restaurant more than five}). It generates a delexicalized response, which can be further lexicalized using the values in the predicted dialog state and the returned entities from the database.

In our implementation, we utilize two separate mT5$_\textit{large}$ models as the backbone for the DST model and the RG model.  As discussed later, we opt for the large model because it demonstrates a substantial performance advantage over its smaller counterpart.
The data preprocessing, including the linearization of dialog state annotations for training, and the evaluation protocol are based on the established implementation of the SOLOIST system~\cite{peng-etal-2021-soloist}. To ensure up-to-date functionality, our implementation is based on the most recent version 4.30 of the Huggingface transformers repository. Our system is designed to prioritize simplicity and efficiency, with the primary goal of minimizing the complexity and effort required for training, evaluation, and future development. We report the standard evaluation metrics for the E2E task, including the Inform Rate, Success Rate, and the delexicalized corpus BLEU score.\footnote{All E2E experiments were run on a single A100 80 GiB GPU; batch size of 4 and a learning rate of $5e-5$; 5 epochs.}

\begin{table}[]
\centering
\def\arraystretch{0.85}
{\footnotesize
\begin{tabularx}{\linewidth}{X lccc@{}}
\toprule
\multirow{2}{*}{\bf Language} &  & \multicolumn{3}{c}{\bf End-to-End Modeling} \\ \cmidrule(l){2-5} 
                          &  & Inform       & Success      & BLEU      \\ \midrule
ENG                       &  & 67.9         & 39.0         & 15.7      \\
ARA                       &  & 66.8         & 36.7         & 14.0      \\
FRA                       &  & 47.9         & 22.2         & 12.0      \\
TUR                       &  & 45.9         & 21.2         & 16.7      \\
AVG.                      &  & 57.1         & 29.8         & 14.6      \\ \bottomrule
\end{tabularx}%
}%
\caption{Fully supervised E2E performance for mT5$_\textit{large}$. AVG. shows the mean average of the evaluation scores across all four languages. The reported scores are averaged over 3 random runs.}
\label{tab:table_e2e}
\vspace{-1mm}
\end{table}

Table \ref{tab:table_e2e} presents the results of the fully supervised E2E experiments. As anticipated, we observe noticeable performance disparities across languages, particularly in comparison to English. Furthermore, we find that the size of the pretrained language model significantly impacts system performance. Specifically, the mT5$_\textit{large}$ model exhibits a substantial (mean average) performance improvement of 16.4 Inform Rate, 17.2 Success Rate, and 4.6 BLEU points, compared to mT5$_\textit{small}$.

\section{Conclusion}
\label{s:conclusion}

We have introduced a large-scale, culturally adapted, multilingual, and multi-parallel training and evaluation framework for \tod, which covers $\sim$495,000 dialog turns over 4 languages. The dataset was motivated by the limitations of current \tod datasets in multilingual setups, which we systematically analyzed as one contribution of this work. Owing to its unique set of properties and scale, beyond initial analyses and experiments conducted in this work, we hope that \dataset will inspire a wide array of further developments in modeling, analysis, and interpretability of multilingual and cross-lingual multi-domain \tod.

For instance, future work could replicate the data collection process to expand the dataset to even more languages (including low-resource ones). Further, one could analyze the performance disparities observed in Tables~\ref{tab:nlu-results}-\ref{tab:table_e2e} within each language-specific \tod system, as well as explore methods to mitigate such disparities, e.g., through the utilization of cross-lingual transfer techniques. Future work could also explore evaluation metrics beyond the ones explored in this work, e.g., it would be interesting to explore the correlation between the increase in evaluation scores in multilingual \tod systems and the resulting performance gain in terms of factors such as utility, user experience, and user satisfaction.  Additionally, it would be important to investigate how \tod systems should, ideally, be constructed and evaluated across different languages to ensure their inclusiveness and robustness in diverse linguistic contexts. 

\rparagraph{Code and Data}
We release the dataset and code at \href{https://github.com/cambridgeltl/multi3woz}{\nolinkurl{github.com/cambridgeltl/multi3woz}}.

\section*{Acknowledgments}

Songbo Hu is supported by Cambridge International Scholarship. Ivan Vuli\'{c} acknowledges the support of a personal Royal Society University Research Fellowship (no 221137; 2022--).

We would like to thank our internship students, Bassil Alaeddin (for the work on the Arabic portion of the dataset) and Max Letellier (for French), for their contributions and dedication to this project. We are grateful to a large number of our diligent annotators for their significant efforts and contributions to this work. Furthermore, we would like to express our gratitude to the TACL editors and anonymous reviewers for their insightful feedback, which greatly improved the quality of this paper.

\clearpage
\bibliography{references}
\bibliographystyle{acl_natbib}

\end{document}